\documentclass{article}

\usepackage{arxiv}
\usepackage{booktabs} 
\usepackage{caption}  
\usepackage{graphicx,color,psfrag}
\usepackage{url}
\usepackage{amsmath}

\graphicspath{{./}}

\usepackage[utf8]{inputenc} 
\usepackage{geometry} 
\usepackage{booktabs} 
\usepackage{array} 

\definecolor{mygray}{gray}{0.9}

\usepackage[backend=biber]{biblatex}
\bibliography{Administrative Data}
\bibliography{Signal Processing}
\bibliography{Health}
\bibliography{Homelessness AI}
\bibliography{Homelessness Patterns}
\bibliography{ML Fundamentals}
\bibliography{Sequences and Sessions}
\bibliography{Musa_References}

\newcommand{\bE}{\begin{enumerate}}
\newcommand{\eE}{\end{enumerate}}
\newcommand{\bI}{\begin{itemize}}
\newcommand{\eI}{\end{itemize}}

\AtEveryBibitem{\clearfield{isbn}}
\AtEveryBibitem{\clearfield{issn}}

\begin{document}
\sloppy

\title{Efficient Observation Time Window Segmentation for Administrative Data Machine Learning}

\author{Musa Taib and Geoffrey G.~Messier  \\
  University of Calgary  \\
  2500 University Dr.~NW, Calgary, AB, Canada, T2N 1N4\\
  gmessier@ucalgary.ca
}

\date{}

\maketitle

\begin{abstract}
Machine learning models benefit when allowed to learn from temporal trends in time-stamped administrative data.  These trends can be represented by dividing a model's observation window into time segments or bins.  Model training time and performance can be improved by representing each feature with a different time resolution.  However, this causes the time bin size hyperparameter search space to grow exponentially with the number of features.  The contribution of this paper is to propose a computationally efficient time series analysis to investigate binning (TAIB) technique that determines which subset of data features benefit the most from time bin size hyperparameter tuning.  This technique is demonstrated using hospital and housing/homelessness administrative data sets.  The results show that TAIB leads to models that are not only more efficient to train but can perform better than models that default to representing all features with the same time bin size.
\end{abstract}

\keywords{administrative data, time window segmentation, machine learning, hospital, homelessness}

\section{Introduction}
\label{sec:Intro}

Utilizing data features from administrative data to predict outcomes for individuals or groups is an exciting application of machine learning.  Administrative data typically consist of records collected while delivering services that can include the justice system, filing taxes, progressing through the education system or accessing housing and homelessness services \cite{lyon2015, young2016, figlio2016, kuhn1998}.  However, with the extensive use of electronic medical records (EMRs), it is arguable that healthcare is the one of the most important sectors to make use of administrative data  \cite{mazzali2015}.

Since most administrative data entries are time stamped, administrative data are fundamentally time series data.  The time dimension carries important trend information for a machine learning model like whether a patient's condition in hospital is improving or getting worse with time.  The length of time between events is also important since there is clearly a difference between someone who accesses an emergency housing shelter seven times in one week versus seven times in one year.  The importance of the time dimension is not lost on most machine learning researchers.  For example, the excellent survey by Morid, et. al. \cite{morid2023} summarizes over 76 papers applying machine learning to medical administrative data and distinguishes them in part by how they account for the temporal nature of that data.

While a popular approach is to simply use temporal information to organize data features into an ordered sequence, understanding when an event does not occur can be as valuable as representing as when it does \cite{lipton2016}.  As a result, a large number of studies represent temporal administrative data using the {\em temporal matrix} approach \cite{morid2023}.  In a temporal matrix, columns correspond to specific events or data features and rows correspond to regularly spaced time intervals within the observation window.  These time intervals are referred to as {\em time bins}.

Time bin duration or ``size'' is an important temporal matrix parameter.  While the simplistic approach is to use the same bin size for all data features, it has been demonstrated that performance and model complexity can be improved when representing different features using different time resolutions \cite{svenson2020, barzegar2021}.  However, the challenge is that allowing each feature to have its own bin size means that the model's hyperparameter search space will grow exponentially with the number of data features.

The contribution of this paper is to present a pre-processing technique that determines which data features benefit the most from time bin size tuning and which features can be represented using a single time bin.  Using dynamic time warping \cite{sakoe1978}, features are ranked according to a new metric that quantifies how much their power for discrimininating between data set labels improves as they are represented at a higher time resolution.   A handful of features from the top of this list are selected for time bin tuning and the remaining features are represented using a single time bin.  Using data sets from the medical and housing/homelessness domains, we will demonstrate that this approach results in simpler models that are quicker to train while still performing as well or better than when all data features are represented using the same time resolution.

\section{Related Work}
\label{sec:Related}

There is a wide variation in how the temporal matrix framework is applied in practice.  A number of studies use a single time bin where the start of the window coincides with a patient first appearing in the data \cite{el-rashidy2020, purushotham2018, harutyunyan2019}.  A single time bin fits within the temporal matrix framework but clearly does not allow the machine learning model to benefit from the trends that appear in data features over time.  Several other studies use multiple time bins and apply the same bin size to all data features but do not justify their choice of bin size \cite{lee2015, lipton2016, harutyunyan2019, zhang2018a, thorsen-meyer2020, min2019}.  

There has been some work on demonstrating how the temporal matrix time bin size affects machine learning model performance.  Varying the size of a single bin that starts from time zero has been investigated in \cite{el-rashidy2020, purushotham2018}.  Min, et. al. varies the size of multiple uniformly sized time bins and demonstrates improved performance with larger bins \cite{min2019}.  

A limited number of studies have also investigated using different bin sizes for different features.  Svenson, et. al. investigate different time bin resolutions and demonstrate that not all data features require the same bin size in order to maximize model performance \cite{svenson2020}.  Barzegar, et. al. take a multi-rate signal processing approach where they determine how best to represent different data features using different sample rates \cite{barzegar2021}.

Most machine learning models can accomodate data features represented using different time resolutions.  Classical machine learning models and non-recurrent neural networks simply treat each time bin of each data feature as a separate data feature input.  Nothing in the machine learning model structure requires that these bins are the same size.  The traditional forms of recurrent neural network (RNN's) models do assume that all data features are represented as a multi-dimensional input with a common time resolution.  However, RNN variations that include ensemble RNNs \cite{barzegar2021} and channelized long short-term memory (LSTM) networks \cite{harutyunyan2019} have been proposed that allow for different features to be represented with different time resolutions.

Time bin size must be selected with care and not all data features need to be represented using the same time resolution \cite{svenson2020, barzegar2021}.  Choosing a common time bin size that is uncessarily small for some data features results in an overly complex model.  This complexity risks data overfitting and can be resource intensive to train \cite{hastie2009}.  However, the size of an exhaustive search for a unique bin size for each data feature grows exponentially with the number of data features and is impractical for many applications.

To the authors' knowledge, only \cite{barzegar2021} has proposed a non-exhaustive search approach for selecting different time resolutions for each feature.  However, the method proposed by \cite{barzegar2021} is based on physics informed machine learning for vehicle sensing applications and is not suitable for general administrative data applications.

This paper addresses this gap by proposing a pre-processing technique called time-series analysis to investigate binning (TAIB) that is suitable for generic administrative data applications.   TAIB produces a ranking of data features in decreasing order of how much the machine learning model will benefit from representing that feature using multiple time bins with increasing resolution.  TAIB makes use of the dynamic time warping (DTW) technique \cite{sakoe1978} that has been used in many applications for robustly comparing the difference between time sequences.

\section{Data Sets}
\label{sec:Data}

The first administrative data set employed in this paper is MIMIC-III (Medical Information Mart for Intensive Care III), a comprehensive and publicly available database widely used in critical care research \cite{johnson2016}. Our objective with the MIMIC-III data set is to predict patient mortality in the intensive care unit (ICU), a significant challenge in medical analytics \cite{liu2022predicting,sadeghiEarlyHospitalMortalityPrediction}. This task is formulated as a binary classification problem, where we use data from the first 48 hours of a patient's ICU stay to predict mortality.

The MIMIC-III database encompasses a wide array of healthcare information across 26 tables, including detailed records of patient demographics, vital signs, laboratory test results, and more. For our mortality prediction analysis, we particularly focus on the CHARTEVENTS and LABEVENTS tables. These tables are crucial as they contain physiological metrics and lab results that are integral to assessing a patient's condition.  As in \cite{purushotham2018}, the specific features selected are those related to the SAPS II (Simplified Acute Physiology Score II) score \cite{legall1993}. SAPS II is a globally recognized severity-of-disease classification system, designed to evaluate the risk of ICU patient mortality.  The selected features include age, vital sign information (heart rate, systolic and diastolic blood pressure, and body temperature) and laboratory results (blood urea nitrogen levels, white blood cell count, potassium levels, sodium levels, bicarbonate levels, bilirubin levels, and Glasgow Coma Scale).  Consistent with \cite{purushotham2018}, the MIMIC-III data is filtered to include only first ICU admissions. We exclude individuals who left the ICU within the first 48 hours and those under 15 years old. 

The second administrative data set contains the records of people utilizing services at the Calgary Drop-In Centre (DI), a large North American emergency housing shelter.  
In addition to a timestamp and scrambled client ID number, this data set includes a categorical entry type field with values outlined in Table~\ref{tb.DiEntryTypes}.  

\begin{table}[htbp]
\centering
\begin{tabular}{cl}
Entry Type & Notes \\ \hline
{\tt Bar} &  Temporary restriction from accessing services.\\
{\tt CounsellorsNotes} &  Consultation with support counsellor.\\
{\tt ProgressDetails} &  Note of progress towards a client's goals.\\
{\tt Log} &  General log entry.\\
{\tt Sleep} &  Check-in to sleep in shelter.\\
{\tt BuildingCheckIn} &  Check-in to building (for any reason).\\
{\tt Mail} &  Letter ready for client pick-up.\\
{\tt Message} &  Message for client.\\
{\tt CondEntry} &  Conditional entry granted to shelter.\\
\end{tabular}
\caption{DI data set record entry types.}
\label{tb.DiEntryTypes}
\end{table}

The DI data set also contains free text case notes associated with most entries.  To protect client privacy, these fields were replaced with keyword count data features by DI staff prior to the release of the data.  The keyword categories, the number of search keywords in each category and some examples of those search words are provided in in Table~\ref{tb.DiKeywords}.  For example, if a record contained the free text comment string ``The client was involved in a fight and was punched.  Police were called.'', that comment field would be replaced by keyword counts of 2 for the Conflict category, 1 for Police/Justice and 0 for all other categories.

\begin{table}[htbp]
\centering
\begin{tabular}{ccc}
Category & Number of Keywords & Example Keywords \\ \hline
Addiction & 151 & alcohol, booze, meth, naloxone \\
Mental Health & 127 & anxious, depressed, trauma, breakdown \\
Conflict & 355 & fight, punch, kick, assaulted \\
Police/Justice & 121 & police, lawyer, incarcerated, arrested \\
Emergency Medical Services (EMS) &  6 & ambulance, EMS, paramedic, medics \\
Supports & 335 & education, housing, counsellor, employment 
\end{tabular}
\caption{DI data set comment field keyword categories.}
\label{tb.DiKeywords}
\end{table}

Clients with any records that occur before July 1, 2008 were removed from the data due to the presence of missing entries and discrepancies in the data prior to that date.  Records for individuals who began using DI services after January 20, 2018 were also discarded to mitigate right truncation of the data.

Following established practice for homelessness administrative data analysis \cite{kuhn1998}, a clustering algorithm is applied to the historical shelter use patterns of people present in the DI data to label each person as having a chronic, episodic or transitional pattern of shelter use.  Chronic shelter use corresponds to long term, sustained shelter use often over many years.  Episodic patterns are also longer term but much more sporadic.  Finally, the transitional label is assigned to people who use shelter for only a short period of time.  Chronic shelter users are typically a minority of the people accessing emergency shelter but their sustained use means they consume the majority of system resources \cite{kuhn1998}.  As a result, they are often prioritized for supportive housing programs.

Our objective with the DI data is to use the first 90 days of a person's record of shelter use to predict whether that person will become a chronic emergency shelter user (the positive case) or an episodic or transitional user (the negative case).  Cluster analysis is typically only performed on historical data to determine a person's shelter use patterns in retrospect.  A machine learning method that uses a person's first 90 days in shelter to predict if they are at risk of becoming a chronic shelter user could be used in real time to connect that person to support programs.

\section{Methods}
\label{sec:Methods}

\newcommand{\Dall}{D_{\rm All}}

\subsection{Data Representation}
\label{ssec:Rep}

Assume a data set containing records for $N$ people.  Each person is associated with one or more time stamped records and each record contains $K$ data features in addition to the time stamp and person identification field.  Following \cite{morid2023}, this data can be represented by $N$ {\em temporal matrices}, each matrix representing the experience of one person.  While each person will typically have a different number of records, the $N$ temporal matrices can be made the same size by defining an observation window of length $T_O$ relative to each person's earliest record.  The observation window can then be divided into $L = T_O/T_B$ equally spaced time bins, where $T_B$ is the duration of each bin.  Bins where no event occurs for a particular person can be filled with zeros or null values.  The data set therefore contains $N$ instances where each instance is an $L \times K$ temporal matrix associated with one person.

Categorical data features are assumed to be one hot encoded so that each categorical value is a separate data feature.  A temporal matrix cell for one of these categorical features contains the number of times that categorical value occurs during the cell's corresponding time bin.  Continuous data features are represented in this paper by taking the average of all the continuous values that fall into each bin.  In general, this averaging operation could be replaced by any mathematical operation on the continuous values.  Continuous value data feature bins that contain no time stamped values are assigned a null value.   Static data features are copied across all time bins.

As noted in Section~\ref{sec:Related}, traditional recurrent neural networks are designed for multi-dimensional data features that occur at regularly spaced time intervals.  This is consistent with the temporal matrix representation described above.  To accomodate different features with different time resolutions, temporal matrix representation can be generalized.  

\begin{figure}[htbp]
\centerline{\includegraphics[width=3in]{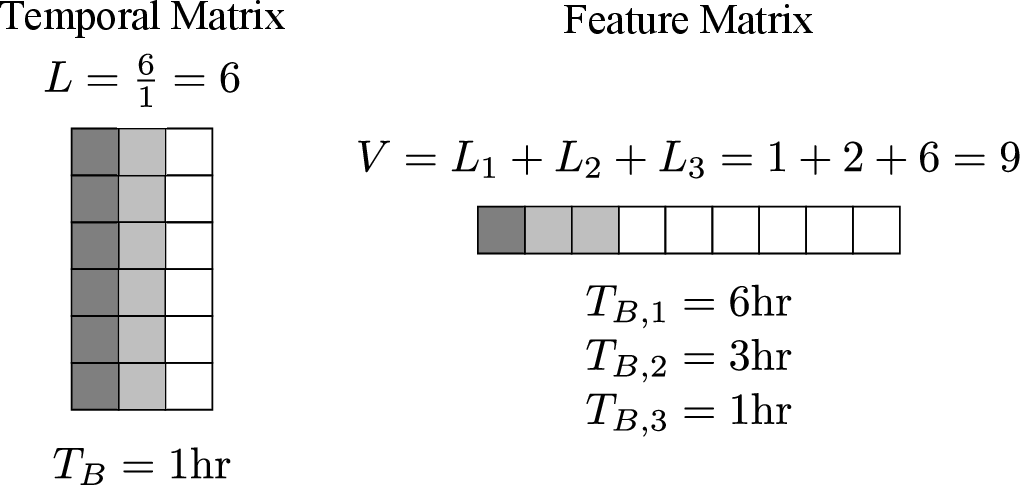}}
\caption{Temporal and feature matrix, single user, $K=3$, $T_O$ = 6 hr.}
\label{fg.FeatureMtx}
\end{figure}

To elaborate, let the notation above be generalized so that $T_{B,k}$ is the bin size for data feature $k$.  This allows for the possibility that $T_{B,k} \ne T_{B,r}$ for $k \ne r$.  The number of bins used to represent feature $k$ is $L_k = T_O/T_{B,k}$.  The entire data set can now be represented as an $N \times V$ {\em feature matrix} where each row represents all of the data associated with a single person.  Specifically, each row contains  $K$ concatenated feature vectors of length $L_k$, $k=1\ldots K$ so that $V = \sum_{k=1}^K L_k$.  A simple example is shown in Figure~\ref{fg.FeatureMtx}.  This figure shows how the data for a single person would be represented as a temporal matrix and as a single row in a feature matrix.  Note that the temporal matrix is restricted to using the same bin size (same number of rows) for each feature but the feature vector can choose a different sized bins.

The feature matrix format can be used by any machine learning algorithm not designed specifically for regularly spaced temporal values.  These algorithms include all of the ``classic'' machine learning models (ie. logistic regression, variations on the decision tree, support vector machines) \cite{hastie2009} as well as multi-layer perceptron and convolutional neural networks \cite{Goodfellow-et-al-2016}.  These algorithms do not depend on a common temporal bin size since they treat each of the $V$ values in a feature matrix row as distinct data features.  However, these algorithms can still benefit from temporal trends that help to differentiate people.  For example, a person accessing an emergency room sporadically will have a different pattern of values in their feature vector than someone else who accesses an emergency room regularly.  Any machine learning algorithm benefits from different data feature patterns, even if the algorithm isn't explicitly designed to recognize that the patterns vary over time.

\subsection{Efficiently Assigning Bin Size}
\label{ssec:Taib}


The size of the time bins used in the temporal matrix and feature matrix representations described in Section~\ref{ssec:Rep} is an important design parameter.  At one extreme, a single bin (ie. $T_{B,k} = T_O$, $k=1\ldots K$) results in both the temporal and feature matrix representations reducing to $N$ vectors of length $K$.  This reduces data set size and decreases model training time.  However, it also discards all temporal information which degrades model performance for many data sets.  At the other extreme, $T_{B,k}$ could be set to the maximum resolution available, often 1 second for many health data sets.  While this captures all temporal trends in the data, this may also result with a large and very sparse data set with many null entries.  The penalty is greatly increased training time and degraded performance due to the curse of dimensionality \cite{hastie2009}.

This suggests that an optimal bin size that maximizes model performance may exist between these two extremes.  For the temporal matrix representation this optimal bin size can be discovered for most data sets using a brute force sweep of $T_B$ over a range of reasonable values.  However, for data sets with very different data features, enforcing a single bin size may result in too much time resolution for some features and not enough for others.  While feature matrix representation solves this problem by allowing each of the $K$ features to have its own value of $T_{B,k}$, the search space to find the optimal bin size for each data feature grows exponentially with $K$.

This paper proposes the Time Series Analysis to Investigate Binning (TAIB) method for ranking data features based on which feature has the potential to benefit most from being represented using higher resolution time bins.  Assuming a binary classification problem, the method begins by separating the instances of feature $k$ into positive and negative groups using the binary outcome label.  TAIB is based on the notion that feature $k$ will benefit from representation by smaller time bins if the positive and negative groups start to appear more ``different'' as $L_k$ increases.  Features with groups that appear more or less the ``same'' regardless of $L_k$ could be represented with fewer or even a single bin.

\newcommand{\Nset}{\mathcal{N}}
\newcommand{\Pset}{\mathcal{P}}
\newcommand{\CntN}{C_\mathcal{N}}
\newcommand{\CntP}{C_\mathcal{P}}
\newcommand{\Lset}{\mathcal{L}}

Specifically, begin by defining the simply ordered set $\Lset$ as containing a range of possible bin size values.  For feature $k$ and a particular value $L_k$ from $\Lset$, normalize across instances and across time so that the feature vector values have zero mean and unit variance.  The feature vectors are then divided based on binary outcome label into two disjoint groups of vectors $\Nset$ and $\Pset$ so that $|\Nset|+|\Pset|=N$, where $|\cdot|$ indicates the cardinality of a set.  Let the centroid of the positive and negative groups be represented as $\CntP$ and $\CntN$, respectively.  The separation or ``difference'' between the positive and negative feature groups for $L_k$ time bins is then calculated according to

\begin{equation}
S_{k,L_{k}} = 
  \frac{1}{|\Nset|} \sum_{n \in \Nset} {\rm DTW}\left(n,\CntP\right)
+ \frac{1}{|\Pset|} \sum_{p \in \Pset} {\rm DTW}\left(p,\CntN\right)
\label{eq.DtwDist}
\end{equation}

\noindent
where ${\rm DTW}(a,b)$ is the dynamic time warped Euclidean distance between vectors $a$ and $b$ \cite{sakoe1978}.  This process is repeated for each value of $L_k$ in the simply ordered set $\Lset$.  The corresponding values of $S_{k,L_k}$ calculated for each value in $\Lset$ form an $|\Lset|$ element vector $\mathbf{S}_k$.

Dynamic time warping (DTW) \cite{sakoe1978dynamic} warps the time axes of two vectors $a$ and $b$ in order to achieve the maximum co-incidence between them \cite{han2011data}.  Since the objective is to spot when adjusting $L_k$ results in a noticeable difference between features, DTW can be seen as a conservative choice since the warping is an attempt to minimize the distance between two feature vectors.  This means that when a difference between vectors is observed, using DTW distance allows us to be more confident it is a meaningful change caused by adjusting $L_k$ and not a random artifact of the data.

A feature $k$ will benefit from smaller time bins if an increasing trend is observed in the values of $\mathbf{S}_k$ with increasing $L_k$. This trend is calculated by determining the line of best fit for the values of $\mathbf{S}_k$ versus $L_k$.  The slope of the best fit line associated with each of the $K$ features is then used to rank them in descending order (ie. the top feature has the largest positive slope) so that features closer to the top of the list benefit more from a finer time resolution.

\subsection{Feature Set Creation}
\label{ssec:FeatureSets}

In general, the framework described in Section~\ref{ssec:Rep} allows each feature to have a unique bin size. 
 However, the results in this paper will simplify this somewhat by dividing features into two groups, $F_1$ and $F_L$.  The features in the first group are assigned a single time bin, $T_{B,k}=T_O, L_k = 1\ \forall\ k \in F_1$.  The features in the second group are assigned $L$ time bins, $T_{B,k} = T_B, L_k = L = T_O/T_B\ \forall\ k \in F_L$ where $T_B \leq T_O$.

The TAIB algorithm described in Section~\ref{ssec:Taib} is used to rank the data features in descending order of how much the feature's discriminative powers would benefit from increased time resolution.  A data set $D_w$ is an $N \times V$ feature matrix created by assigning the top $w$ TAIB features to $F_L$ and the remaining $K-w$ features to $F_1$.   This means that the number of columns in $D_w$ is $V = wL + (K-w)$.  In Section~\ref{sec:Results}, results will be generated using data sets $D_1$ and $D_3$ which are created using the top TAIB feature and top three TAIB features, respectively.  Results will also be generated for $D_{\rm All} = D_K$ which represents the temporal matrix scenario where all features are placed in $F_L$ and share a common time bin size.  

As defined above, the feature matrix data sets have only one parameter, $L$, that must be swept to determine optimal bin size.  The number of bins will be varied from 1 to 90.  This means the maximum time resolution for the DI data set will be $T_O/L = {\rm 90\ days}/90 = 1$ day and the maximum resolution for the MIMIC data set will be $T_O/L = {\rm 48\ hours}/90 = 32$ minutes.  Tables ~\ref{tb.DataSetsLeft} and ~\ref{tb.DataSetsRight} demonstrate how increasing $L$ affects both data set size and sparsity (percentage of empty data elements) for $\Dall$ and $D_3$.  Note that these tables are calculated after applying one-hot encoding to any categorical variable described in Section~\ref{sec:Data}.  

Tables \ref{tb.DataSetsLeft} and \ref{tb.DataSetsRight} indicate a high degree of sparsity even for $L=1$.  This is primarily because of the sparse nature of one-hot encoding.  However, the sparsity still increases considerably with increasing $L$ indicating that higher time resolution also has an impact.  Comparing Table~\ref{tb.DataSetsLeft} with Table~\ref{tb.DataSetsRight} also demonstrates that using TAIB to select only the top 3 features for a higher time resolution significantly reduces the number of features ($V$) that the machine learning model must contend with.

After the data sets are created, they are divided into three sets train/validation/test in the ratio 70\%/15\%/15\%. The TAIB algorithm and the ML models were both trained and validated using the first two sets while the results for the model's capabilities were tested using only the test set. Moreover, to address the potential impact of random initialization on the performance of the machine learning models, each model underwent 40 separate training and testing cycles. Each cycle involved re-initializing the model's parameters randomly before training it again with the training set and validating it with the validation set. This process helps in assessing the consistency and reliability of the models’ performance, as it accounts for variability that might arise due to different starting conditions in the learning process.

\begin{table}[htbp]
\centering
\begin{minipage}[t]{0.48\textwidth}
\centering
\begin{tabular}{c||cc|cc}
  & \multicolumn{2}{c}{DI} & \multicolumn{2}{c}{MIMIC} \\
$L$ & $V$ & Percent Empty & $V$ & Percent Empty \\ \hline
1 & 23 & 76.26 & 48 & 76.182 \\
90 & 2070 & 86.10 & 4320 &  89.291 \\
\end{tabular}
\captionof{table}{Size and sparsity of $D_{\text{All}}$}
\label{tb.DataSetsLeft}
\end{minipage}\hfill
\begin{minipage}[t]{0.48\textwidth}
\centering
\begin{tabular}{c||cc|cc}
  & \multicolumn{2}{c}{DI} & \multicolumn{2}{c}{MIMIC} \\
$L$ & $V$ & Percent Empty & $V$ & Percent Empty \\ \hline
1 & 23 & 76.02 & 48 & 76.07 \\
90 & 290 & 86.10 & 315 & 78.87 \\
\end{tabular}
\captionof{table}{Size and sparsity of $D_3$}
\label{tb.DataSetsRight}
\end{minipage}
\end{table}

\subsection{Algorithms}
\label{ssec:Algorithms}

The results of varying time bin size and applying TAIB for high time resolution feature selection are presented in Section~\ref{sec:Results} for the following machine learning algorithms.

\subsubsection{Classic Models}\ \\
\textbf{Logistic Regression:} Logistic regression is selected as a simple baseline algorithm and due to its popularity for healthcare applications \cite{cox1958}. The version of logistic regression used in this paper is from Sci-kit Learn with class weights inversely proportional to class frequencies due to the imbalance nature of the data \cite{scikit-learn-logistic-regression}.  The solver defaults to the limited memory variant of the BFGS algorithm with a regularization strength of 1.0.

\textbf{Random Forest:} Random Forest is selected since it is robust against overfitting and adept at handling a large number of features \cite{groll2018prediction}. Implementation utilizes Sci-kit Learn with class weights inversely proportional to class frequencies \cite{scikit-learn-random-forest}.  The number of trees in the forest is 300 and the maximum number of features is set to the root of all the features submitted to the model. The minimum samples per leaf is set to 1 and the minimum samples per split is 2. The max depth per tree was set to 30.

\subsubsection{Deep Learning Models}

Deep Learning models leverage neural networks to extract complex patterns from large data sets \cite{LeCun2015} and are considered due to the superior performance they offer for a variety of applications. In this study, the DL models employ the ReLU activation \cite{Glorot2011} function for hidden layers to mitigate the vanishing gradient problem and a sigmoid \cite{Goodfellow-et-al-2016} function in the final layer for binary classification. The Adam optimizer \cite{Kingma2014} with varying learning rates is used for its efficiency in large-scale applications. Dropout regularization is applied to each dense layer (except the final one) to prevent overfitting, with dropout \cite{Srivastava2014} rates varying between 0.1 and 0.4. The models use binary cross-entropy as the loss function and mini-batch gradient descent with different batch sizes for optimization \cite{Goodfellow-et-al-2016}.

\textbf{Convolutional Neural Networks (CNNs):} CNNs, traditionally used in image processing, have been adapted for time-series data analysis \cite{kashiparekh2019convtimenet}.  As shown in Table \ref{table:4.1}, the CNN utilized in this paper has 1D convolutional layers, followed by batch normalization for stabilized learning, and max pooling layers for dimensionality reduction \cite{Goodfellow-et-al-2016} . This approach allows our CNN model to capture both spatial and temporal patterns in the data. The flattened data is then processed by dense layers, combining spatial hierarchies with classification tasks which makes CNNs particularly effective for analyzing the different temporal time patterns present in the administrative data \cite{Michelucci2019AdvancedAD}.

\textbf{Multi-layer Perceptrons (MLPs):} Multi-layer perceptrons (MLPs) are a fundamental type of neural network made of dense layers that excel at discerning complex patterns within input data  \cite{Goodfellow-et-al-2016}. Their relative simplicity over more complex deep learning structures are useful in interpreting the results presented in Section~\ref{sec:Results}.   Table~\ref{table:4.1} presents three different MLP configurations for a low (small), medium (average) and high (large) complexity model which represent different compromises between performance and the resources required for training \cite{Calin2020DeepLA}.  The effect of MLP model complexity on the impact of time bin size optimization will be demonstrated in Section~\ref{sec:Results}.

\textbf{Gated Recurrent Units (GRUs):} GRUs are a streamlined variant of RNNs that are particularly adept at processing sequential and temporal data. Their architecture is simpler than long short-term memory (LSTM) networks and involves two gates - reset and update - that efficiently modulate the flow of information \cite{cho-etal-2014-properties}. This structure not only allows for more efficient computation but also effectively captures temporal dependencies, crucial for learning from sequences. GRUs have been shown to be particularly effective for administrative data \cite{che2018recurrent}. In this paper, we utilize a traditional GRU structure that utilizes data in temporal matrix form.  The layered structured of the GRU model used in this paper is shown in Table \ref{table:4.1}.

\begin{table}[h!]
\centering
\begin{tabular}{|c|c|c|c|c|}
\hline
\textbf{Model} & \textbf{Stage 1} & \textbf{Stage 2} & \textbf{Stage 3} & \textbf{Stage 4} \\
\hline
\textbf{MLP$_{\text{Large}}$} & Dense(256) & Dense(128) & Dense(32) & Dense(1) \\
\textbf{MLP$_{\text{Average}}$} & Dense(128) & Dense(32) & Dense(1) & N/A \\
\textbf{MLP$_{\text{Small}}$} & Dense(32) & Dense(1) & N/A & N/A \\
\textbf{GRU} & GRU(128) & Dense(64) & Dense(32) & Dense(1) \\
\textbf{CNN} & 1D-CNN(64,3)+BN+MP(2) & 1D-CNN(128,3)+BN+MP(2) & Flatten layer & Dense(64) \\
\hline
\end{tabular}
\caption{Deep learning model parameters.}
\label{table:4.1}
\end{table}

\section{Results}
\label{sec:Results}

\subsection{Framework of the Experiments}
In the following sections, results are presented that demonstrate the impact of varying the number of time bins, $L$, used by the data set $F_L$ described in Section~\ref{ssec:FeatureSets}.  The value of $L$ is varied from 1 to 90 and plotted versus machine learning model F1 scores calculated using the test set split described in Section~\ref{ssec:FeatureSets}.  It is important to note that the observation window (\(T_O\)) was kept constant across all experiments ($T_O$ = 90 days for the DI data set and $T_O$ = 48 hours for the MIMIC data set).  This means that increasing $L$ corresponds to a higher resolution time representation (ie. smaller time bin sizes).  Lastly, each simulation in every experiment was run 40 times with the same data but non-seeded models to ensure the randomness of the initialization of weights have minimal impact on the outcomes of the experiments. 

\subsection{Experiment 1: Model Class and Architecture Variations}
\begin{figure}[htbp]
\centerline{\includegraphics[width=0.75\textwidth]{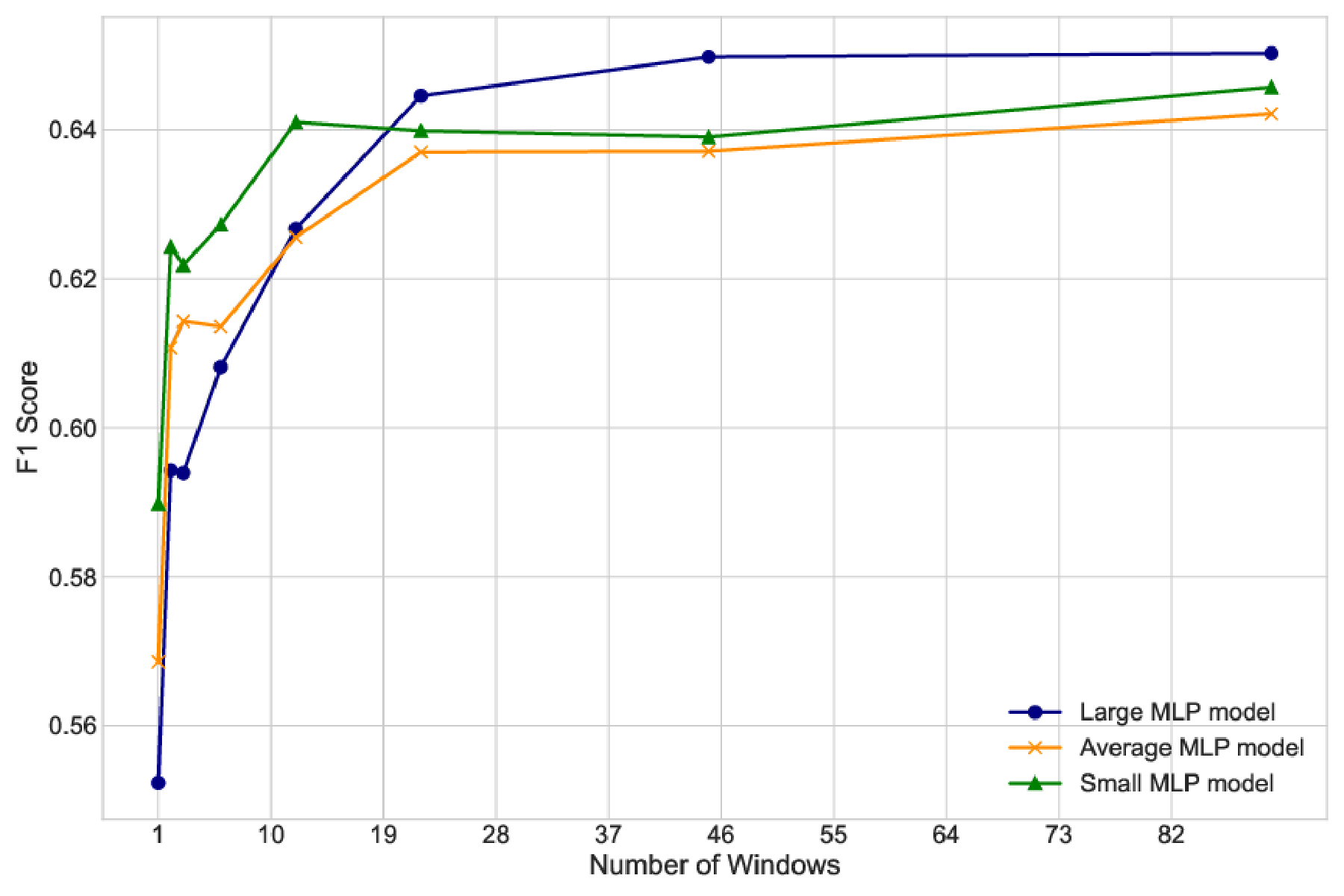}}
\caption{Experiment 1: Chronic shelter use (DI data set).}
\label{fig:Chronic Experiment 1}
\end{figure}

\begin{figure}[htbp]
\centerline{\includegraphics[width=0.75\textwidth]{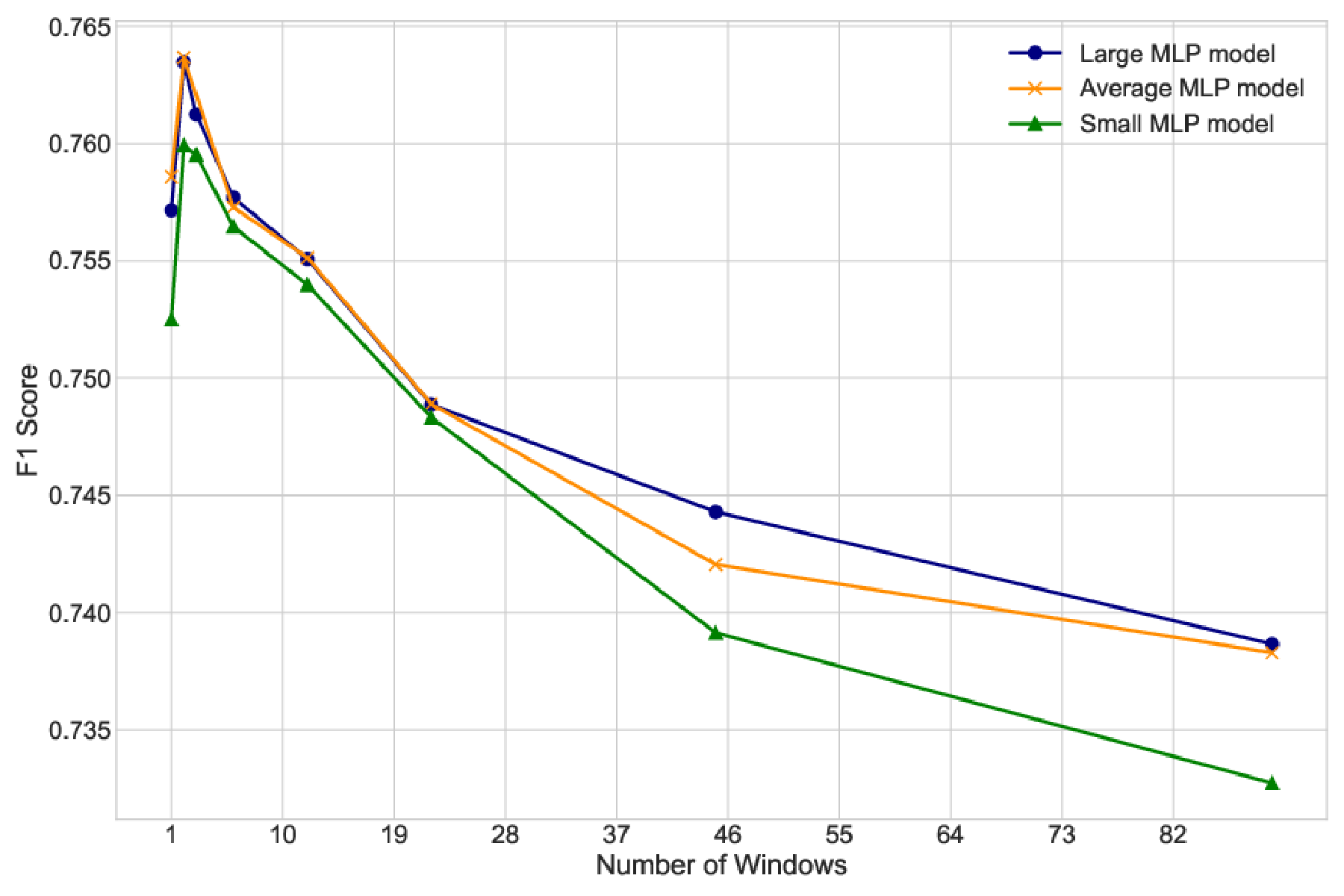}}
\caption{Experiment 1: Patient mortality (MIMIC-III data set).}
\label{fig:Mortality Experiment 1}
\end{figure}

This experiment explores how time bin size affects the performance of the same machine learning model class with different levels of architecture complexity. Figures~\ref{fig:Chronic Experiment 1} and \ref{fig:Mortality Experiment 1} show MLP performance for the small, average and large model architectures described in Table~\ref{table:4.1}. The dataset used for these experiments was $D_{All}$

Figure~\ref{fig:Chronic Experiment 1} demonstrates that window size has a significant impact on how effectively the MLP can solve the chronic shelter use problem.  Performance increases rapidly with $L$ and then reaches a plateau.  As discussed in Section~\ref{sec:Data}, the machine learning model must distinguish between consistent shelter access (chronic) and more sporadic access (episodic and transitional).  While difficult to distinguish with a single time bin ($L=1$), the patterns become much more apparent as $L$ increases.  Performance reaches a plateau in Figure~\ref{fig:Chronic Experiment 1} which suggests the sparsity in the DI data set with increasing $L$ does not degrade performance.  In contrast, the performance in Figure~\ref{fig:Mortality Experiment 1} reaches a clear maximum and then degrades with $L$ and the sparse data prevents the training phase from producing a model that generalizes well to the test data.

While architecture complexity does not significantly alter the nature of the curves in Figure~\ref{fig:Mortality Experiment 1}, Figure~\ref{fig:Chronic Experiment 1} does indicate the simpler models are quicker to improve as $L$ is increased from 1.  While more investigation is necessary, a possible interpretation is that a modest number of time bins is sufficient to see a significant jump in low to medium complexity models. The fact that the large model does not provide the best performance until further along the x-axis suggests that the larger models, with their greater capacity, require more information (ie. more bins) to realize their superior performance, whereas smaller models plateau in performance due to their limited capacity.

\subsection{Experiment 2: Consistent Model Class and Architecture}
\label{ssec:Experiment2}
This experiment compares the machine learning performance achieved using the temporal matrix $D_{\rm All}$ to the performance using feature sets $D_1$ and $D_3$ created using the TAIB algorithm. The average size MLP model from Experiment 1 was used here. The results of performing TAIB ranking on the features in the DI and MIMIC-III data sets are shown in Tables~\ref{table:T_Chronic} and \ref{table:T_Mortality}, respectively.  Data features that represent a one-hot encoded categorical variable value are indicated by ``(1Hot)''.  The data sets $D_1$ use the top feature in these tables and $D_3$ use the top three features in these tables.

\noindent 
\begin{center} 
\begin{minipage}[t]{0.32\textwidth} 
\centering
\begin{tabular}{cc}
\toprule
Feature & Scores \\ 
\midrule
Sleep (1Hot) & 1.8495 \\ 
ProgressDetails (1Hot) & 0.3895 \\ 
EMS & 0.3688 \\ 
Log (1Hot) & 0.3584 \\
Bar (1Hot) & 0.0651 \\
\bottomrule
\end{tabular}
\captionof{table}{Chronic Clients}
\label{table:T_Chronic}
\end{minipage}%
\hspace{4mm} 
\begin{minipage}[t]{0.32\textwidth} 
\centering
\begin{tabular}{cc}
\toprule
Feature & Scores \\ 
\midrule
Blood Pressure & 0.8842 \\ 
Heart Rate & 0.8833 \\ 
Cancer & 0.8268 \\ 
Temperature & 0.5667 \\ 
GSC\_Motor\_1 (1Hot)  & 0.5067 \\ 
\bottomrule
\end{tabular}
\captionof{table}{Mortality Prediction}
\label{table:T_Mortality}
\end{minipage}
\end{center}

Both figures demonstrate that the feature vectors outperform the temporal matrix representation with representing just the top TAIB feature offering the best overall performance for moderate window sizes.  The original motivation of using TAIB to prioritize features for time windowing was to simplify the complexity of training with an unnecessarily large set of features.  However, these results reveal that TAIB can also contribute to an improvement in performance.  Representing a feature with a large number of time bins that it does not require creates an unnecessary amount of data set sparsity.  This unnecessary sparsity is what degrades the performance of $D_{\rm All}$ and $D_3$ relative to $D_1$.

\begin{figure}[htbp]
\centerline{\includegraphics[width=0.75\textwidth]{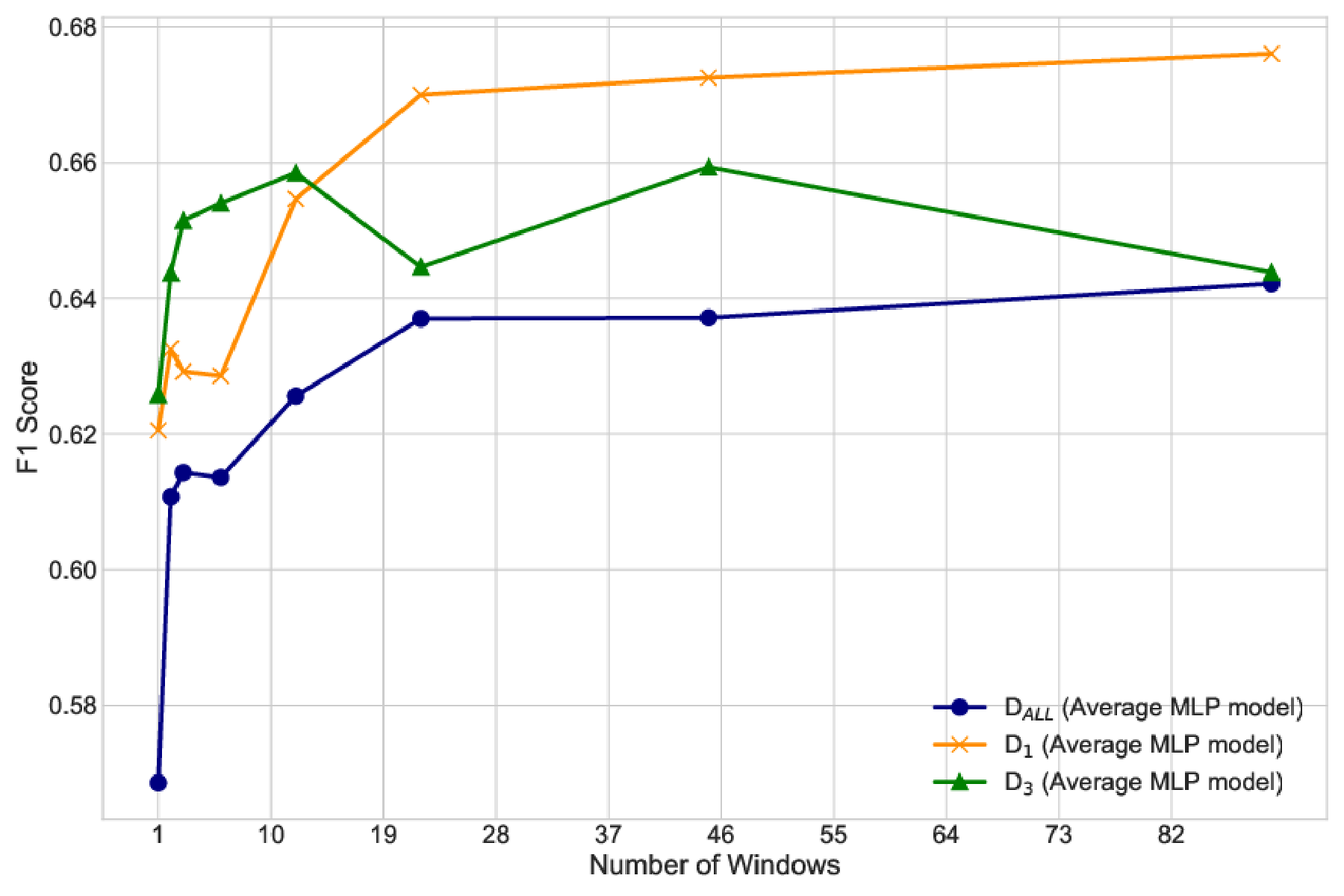}}
\caption{Experiment 2: Chronic shelter use (DI data set).}
\label{fig:Chronic Experiment 2}
\end{figure}

\begin{figure}[htbp]
\centerline{\includegraphics[width=0.75\textwidth]{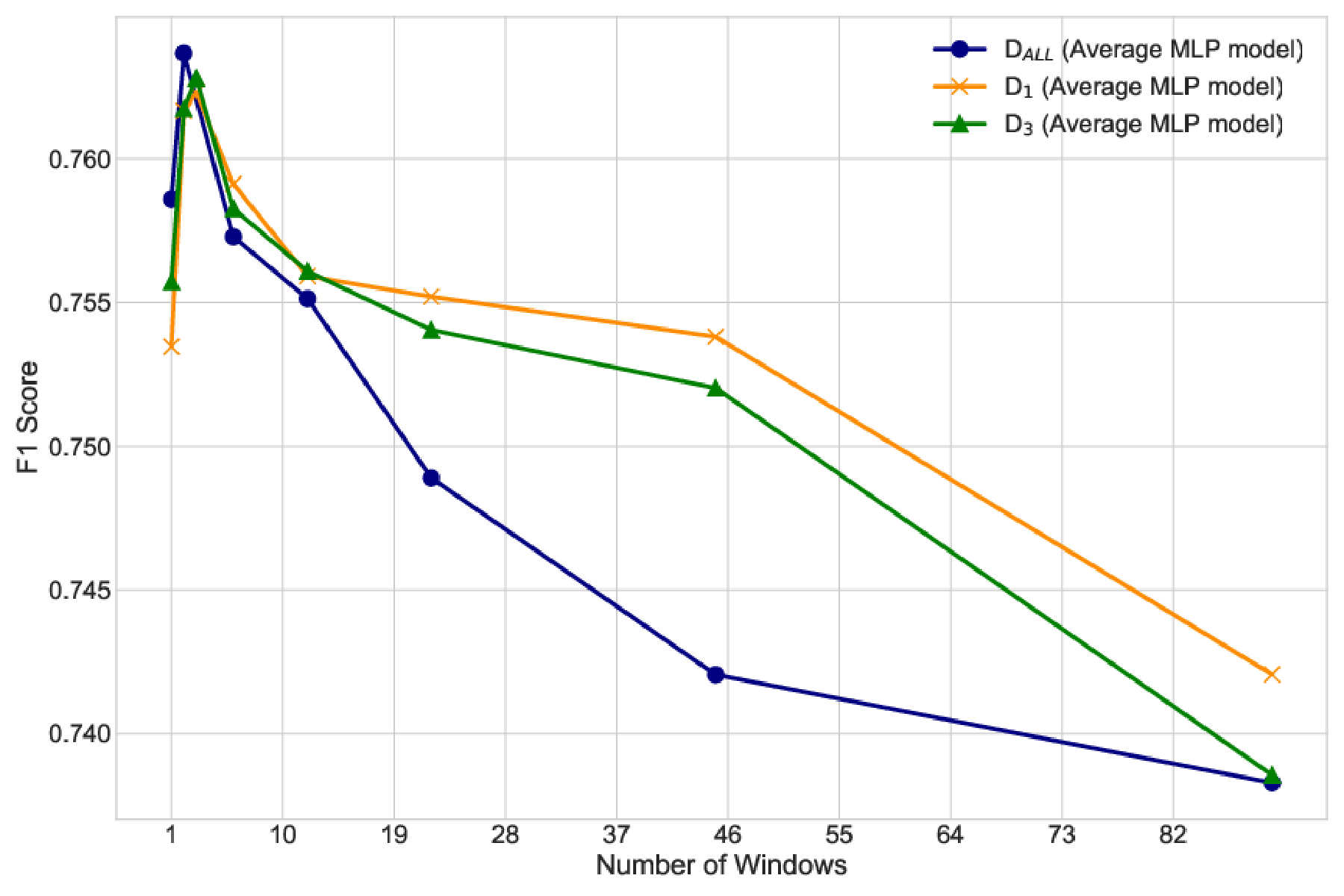}}
\caption{Experiment 2: Patient mortality (MIMIC-III data set).}
\label{fig:Mortality Experiment 2}
\end{figure}

\subsection{Experiment 3: Different Model Classes and Architectures}

\begin{figure}[htbp]
\centerline{\includegraphics[width=0.75\textwidth]{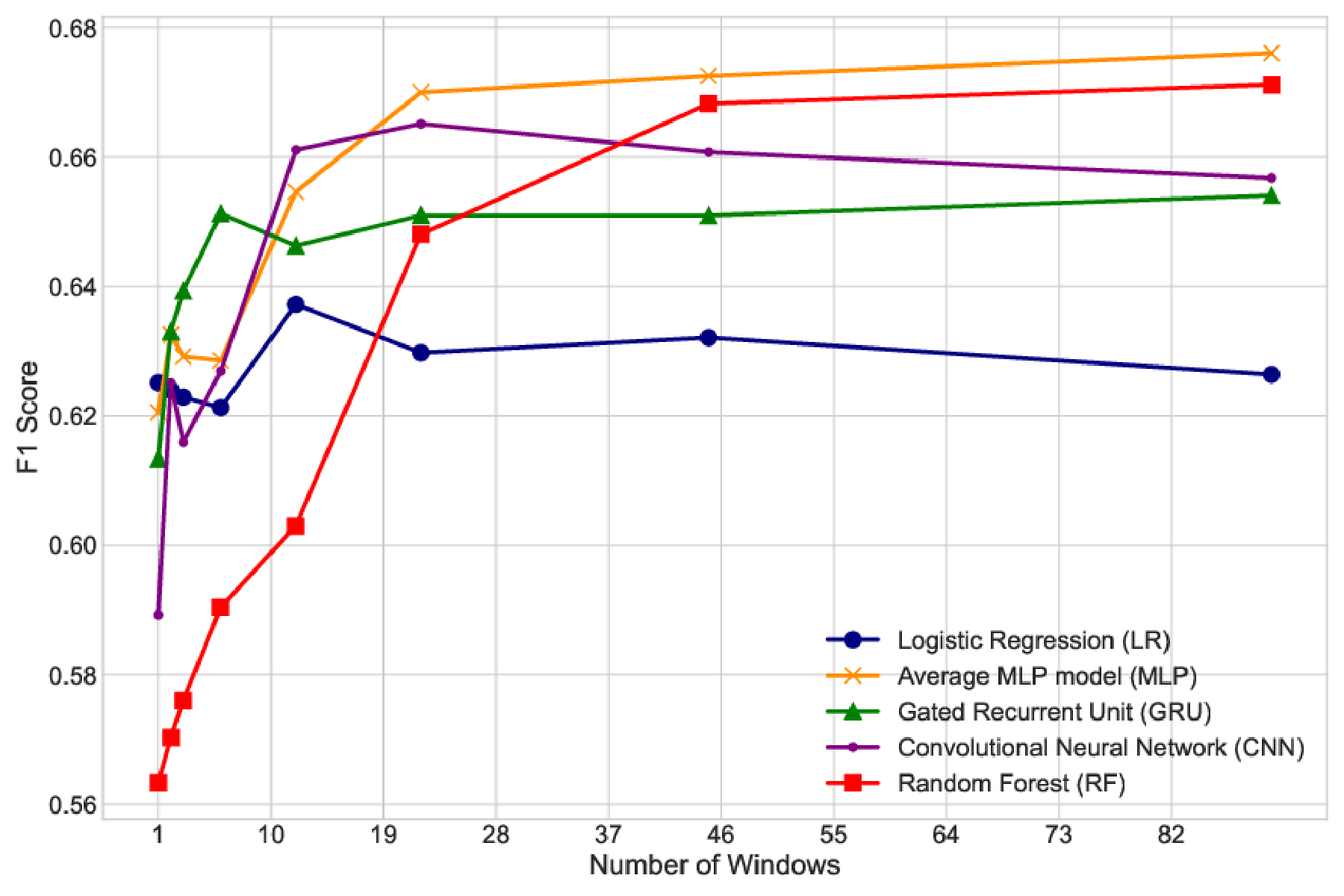}}
\caption{Experiment 3: Chronic shelter use (DI data set).}
\label{fig:Chronic Experiment 3}
\end{figure}

\begin{figure}[htbp]
\centerline{\includegraphics[width=0.75\textwidth]{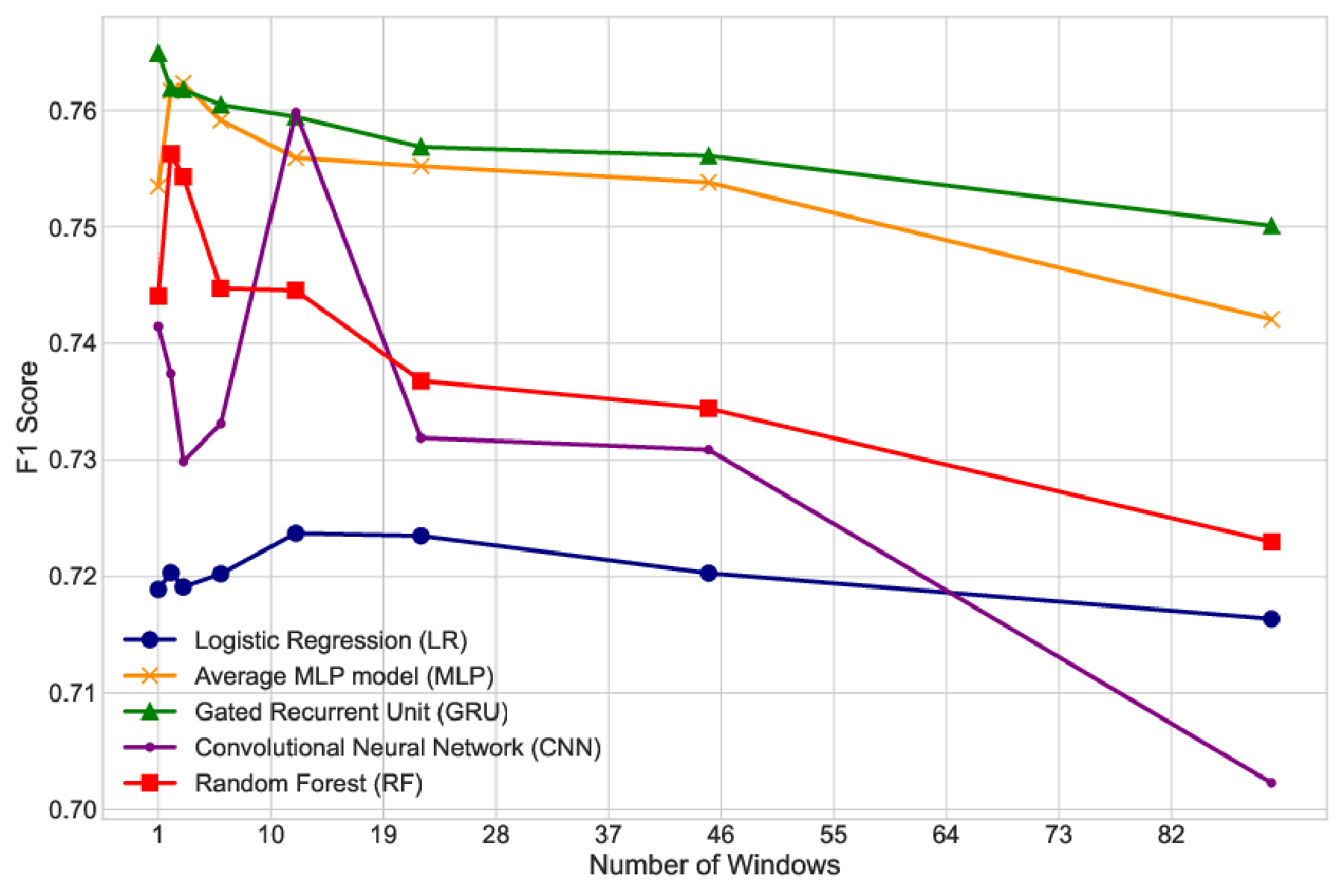}}
\caption{Experiment 3: Patient mortality (MIMIC-III data set).}
\label{fig:Mortality Experiment 3}
\end{figure}

This final experiment compares the performance of all the machine learning models described in Section~\ref{ssec:Algorithms} when operating on $D_{\rm 1}$ except for the GRU for which $D_{\rm ALL}$ is used. The results are shown in Figures~\ref{fig:Chronic Experiment 3} and \ref{fig:Mortality Experiment 3}.  Overall, the relative performance of the algorithms looks as expected with the more complex models offering superior performance.  However, the more important conclusion is that the performance of each algorithm varies considerably with $L$.  This emphasizes that properly choosing time bin size is important regardless of what machine learning algorithm is being employed.

\section{Discussion}
\label{sec:discussion}

\subsection{Assigning Different Time Bins to Features Helps Efficiency and Performance}

Using TAIB to prioritize which features are represented using an increased number of time bins will always improve model training efficiency relative to the temporal matrix representation.  The number of data features, $V$, in Tables \ref{tb.DataSetsLeft} and \ref{tb.DataSetsRight} is far lower (less than half) when using $D_3$ compared with $D_{\rm All}$.  This not only reduces storage requirements but it also accelerates the training process.  This is particularly important for applications with a large number of data features.

Also, as noted in Section~\ref{ssec:Experiment2}, using a feature vector representation informed by the TAIB algorithm may also improve performance.  This is clearly the case in Figure~\ref{fig:Chronic Experiment 2}.  In Figure~\ref{fig:Mortality Experiment 2}, there is very little difference between the maximum temporal matrix performance achieved with $D_{\rm All}$ when compared with the feature vectors.  However, the performance of the feature vectors decline more gradually with increasing $L$ when compared with the temporal matrix results.  This is an advantage in the model tuning process.  If we have confidence that feature vector performance is less sensitive to finding the optimal value of $L$, the tuning process can be simplified by reducing the $L$ value search space.

\subsection{Time Bin Size Importance Depends on the Problem}

One trend that becomes apparent in Section~\ref{sec:Results} is that time bin size has a larger effect on performance for the chronic shelter use problem than it does for the mortality prediction problem.  For example, varying $L$ in Figure~\ref{fig:Mortality Experiment 1} causes F1 score to vary approximately 3\% where the variation in F1 is on the order of 10\% in Figure~\ref{fig:Chronic Experiment 1}.

While TAIB will identify the feature that benefits most from adjusting time bin size, the top features identified TAIB will not always necessarily co-incide with the features that have the greatest predictive power.  This can be illustrated by ranking the features from the DI and MIMIC-III data sets according to their mutual information score with the outcome label. The results are shown in Tables~\ref{table:MI_Chronic} and \ref{table:MI_Mortality}.  In the case of the DI data set, the features shown in Table~\ref{table:T_Chronic} and \ref{table:MI_Chronic} are largely the same.  This means that the feature with the most predictive power also benefits from an increased number of time bins.  However, in the case of MIMIC-III, Tables~\ref{table:T_Mortality} and \ref{table:MI_Mortality} show that the most predictive features and the ones that benefit from time bin optimization are different.  

This means that the mortality results in Figures~\ref{fig:Mortality Experiment 1}, \ref{fig:Mortality Experiment 2} and \ref{fig:Mortality Experiment 3} are generated when optimizing the time bin size of features that do not have the most predictive power.  This highlights the importance of coupling TAIB with some indication of how the features rank in terms of predictive power.

\noindent 
\begin{center} 
\begin{minipage}[t]{0.32\textwidth} 
\centering
\begin{tabular}{cc}
\toprule
Feature & Scores \\ 
\midrule
Sleep (1Hot) & 0.03659 \\ 
Police/Justice & 0.03461 \\ 
Age & 0.01260 \\ 
ProgressDetails (1Hot) & 0.0081 \\ 
EMS & 0.00595 \\
\bottomrule
\end{tabular}
\captionof{table}{Chronic Clients}
\label{table:MI_Chronic}
\end{minipage}%
\hspace{4mm} 
\begin{minipage}[t]{0.32\textwidth} 
\centering
\begin{tabular}{cc}
\toprule
Feature & Scores \\ 
\midrule
Age & 0.1238 \\ 
AdmitType\_2 (1Hot) & 0.0710 \\ 
ServiceType\_3 (1Hot) & 0.0700 \\ 
HeartRate & 0.0579 \\ 
AdmitType\_0 (1Hot) & 0.0442 \\ 
\bottomrule
\end{tabular}
\captionof{table}{Mortality Prediction}
\label{table:MI_Mortality}
\end{minipage}
\end{center}

\section{Conclusion}\label{sec:conclusion}

\subsection{Summary}\label{Summary}

This paper presents the first systematic framework for reducing the observation window time bin size hyperparameter search space that can be applied to general administrative data.  Using administrative data from the healthcare and housing/homelessness domains, our results show that using the TAIB algorithm to properly identify which features benefit from a higher time resolution will greatly simplify machine learning model training and can result in improved performance.  By considering two different classification problems and two different data sets, this paper also demonstrates that not all problems will benefit the same amount from optimizing time bin size.  Variability between administrative data machine learning problems is another reason why a systematic method for optimizing time bin size is important.

\subsection{Limitations and Open Problems}
\label{Limitations and Future Possibilities}

Our research approached windowing as a binary decision, where features with a lower cluster separation score either had windowing applied or not. Future work can explore assigning optimal values of $L_k$ to each feature, allowing for a more tailored approach to windowing.  In particular, features with higher TAIB scores can be tuned using a large number of possible time bin sizes and the number of possible bins sizes could be gradually reduced for features with lower scores. 

Future research could also explore how TAIB could be adapted to multi-dimensional data and integrating alternative distance measures like Mahalanobis \cite{Mahalanobis2008} and correlation-based distances. This could potentially refine the algorithm's performance and offer new perspectives on handling complex, multidimensional datasets.  TAIB could also be combined with a pre-processing method that assesses a feature's predictive power.  This may lead to a single method that will select a feature if it both benefits from a higher time resolution and has sufficient predictive power to influence the overall performance of the machine learning model when its time resolution is optimized.

The conclusions of this paper are also limited to the two administrative data machine learning problems considered.  Additional results applying the TAIB framework to a variety of administrative data machine learning problems will offer a more comprehensive understanding of the method's effectiveness.


\section{Acknowledgments}
\label{sec:ack}

The authors would like to acknowledge the support of Making the Shift, the Calgary Drop-In Centre and the Government of Alberta.  This study is based in part on data provided by Alberta Seniors, Community and Social Services.  The interpretation and conclusions contained herein are those of the researchers and do not necessarily represent the views of the Government of Alberta.  Neither the Government of Alberta nor Alberta Seniors, Community and Social Services express any opinion related to this study.


\printbibliography

\end{document}